\pdfoutput=1
\documentclass[11pt]{article}

\usepackage[final]{acl}

\usepackage{times}
\usepackage{latexsym}

\usepackage[T1]{fontenc}
\usepackage[utf8]{inputenc}

\usepackage{microtype}

\usepackage{inconsolata}

\usepackage{graphicx}
\usepackage{amsmath}
\usepackage{amssymb}
\usepackage{mathtools}
\usepackage{amsthm}
\usepackage{enumitem}
\usepackage{wrapfig}
\usepackage{url}
\usepackage{microtype}      % microtypography
\usepackage{dsfont}
\usepackage{colortbl}
\usepackage{tabulary}
\usepackage{etoolbox}
\usepackage[framemethod=TikZ]{mdframed} % for LLM prompt frames
\definecolor{topcolor}{rgb}{1,0.8,0.8}
\definecolor{secondcolor}{rgb}{1,0.87,0.7}
\definecolor{thirdcolor}{rgb}{1,1,0.8}
\usepackage{multirow}
\usepackage{multicol}
\usepackage{booktabs} % for professional tables
\usepackage{makecell}
\usepackage{mathabx}
\usepackage{enumitem}
\usepackage{listings}
\usepackage{pifont}% http://ctan.org/pkg/pifont
\newcommand{\xmark}{\ding{55}}%
\usepackage{tikz}
\usepackage{pgfplots}
\usepackage[export]{adjustbox}
\usepackage{caption}
\usepackage{subcaption}
\usepackage{listings}
\usepackage{caption}
\usepackage{float}
\usepackage{tcolorbox}
\usepackage{xcolor}
\usepackage{soul}
\DeclareCaptionType{listing}[Figure][List of Listings]
\makeatletter
\AtBeginDocument{%
  \let\c@listing\c@figure % Tie the listing counter to the figure counter
}
\makeatother
\usepackage{tabularx}
\usepackage{colortbl}
\usepackage{tabulary}
\usepackage{etoolbox}
\definecolor{palegreen}{rgb}{0.40, 0.68, 0.40}
\definecolor{aliceblue}{rgb}{0.94, 0.97, 1.00}
\definecolor{honeydew}{rgb}{0.94, 1.00, 0.94}

\definecolor{promptcolor}{RGB}{200, 235, 227}
\definecolor{prompttitlecolor}{RGB}{175, 172, 172}

\usepackage[normalem]{ulem}

\mdfdefinestyle{prompt}{% 
    linecolor =   black,
    linewidth =   0.2pt,
    backgroundcolor =  promptcolor,
    roundcorner     =   1pt,
    font = \small\ttfamily,
    innerleftmargin=1.5pt,
    innerrightmargin=1.5pt
}

\newmdenv[%
    roundcorner=5pt, 
    linecolor =   black,
    linewidth =   1pt,
    font = \small\ttfamily,
    subtitlebackgroundcolor=prompttitlecolor, 
    frametitlebackgroundcolor=prompttitlecolor,
    backgroundcolor=promptcolor, 
    frametitle={Generated caption},
    subtitleaboveskip=0.5\baselineskip,
    subtitlebelowskip=0.5\baselineskip,
    ]{captionenv}

\lstdefinestyle{pythonstyle}{
    language=Python,
    basicstyle=\ttfamily\small, % Font and size
    keywordstyle=\bfseries\color{blue}, % Keywords in bold and blue
    stringstyle=\color{orange}, % Strings in orange
    commentstyle=\itshape\color{gray}, % Comments in italic and gray
    showstringspaces=false, % Don't display spaces in strings
    numbers=left, % Line numbers on the left
    numbersep=4pt, % Distance of numbers from code
    numberstyle=\tiny\color{gray}, % Line number style
    breaklines=true, % Break long lines
    backgroundcolor=\color{LightGray}, % Background color
}

\lstdefinestyle{githubpython}{
    language=Python,
    basicstyle=\ttfamily\small, % Font and size
    keywordstyle=\color{Blue}, % Keywords
    stringstyle=\color{DarkGreen}, % Strings
    commentstyle=\color{Gray}, % Comments
    numberstyle=\tiny\color{Gray}, % Line numbers
    identifierstyle=\color{Black}, % Variable names
    emph={self}, emphstyle=\color{DarkRed}, % Highlight specific words
    showstringspaces=false, % Don't display spaces in strings
    numbers=left, % Line numbers on the left
    numbersep=4pt, % Distance of numbers from code
    frame=single, % Bounding box
    breaklines=true, % Automatic line breaks
    backgroundcolor=\color{LightGray!30}, % Background color
    morekeywords={class,def,return,import,from,as,if,elif,else,for,while,try,except,with,assert}, % Extra keywords
}

\definecolor{Blue}{rgb}{0.1, 0.1, 0.85}       % GitHub's keyword blue
\definecolor{DarkGreen}{rgb}{0.2, 0.6, 0.2}   % GitHub's string green
\definecolor{Gray}{rgb}{0.5, 0.5, 0.5}        % GitHub's comment gray
\definecolor{Black}{rgb}{0, 0, 0}             % Default black for identifiers
\definecolor{DarkRed}{rgb}{0.6, 0.2, 0.2}     % GitHub's "self" red
\definecolor{LightGray}{rgb}{0.95, 0.95, 0.95} % Background light gray

\DeclareMathAlphabet{\pazocal}{OMS}{zplm}{m}{n}

\DeclareMathAlphabet\mathbfcal{OMS}{cmsy}{b}{n}

\title{RiTTA: Modeling Event Relations in Text-to-Audio Generation}

\author{Yuhang He \\
  Microsoft Research \\
  \texttt{yuhanghe@microsoft.com} \\\And
  Yash Jain \\
  Microsoft  \\
  \texttt{yash.jain3599@gmail.com} \\\And
  Xubo Liu \\
  CVSSP, University of Surrey, UK\\
  \texttt{xubo.liu@surrey.ac.uk} \\\AND
  Andrew Markham\\
  CS Department, University of Oxford, UK\\
  \texttt{andrew.markham@cs.ox.ac.uk}\\\And
  Vibhav Vineet \\
  Microsoft Research\\
  \texttt{vibhav.vineet@microsoft.com}}

\begin{document}
\maketitle
\vspace{-3mm}
\begin{abstract}

Existing text-to-audio~(TTA) generation methods have neither systematically explored audio event relation modeling, nor proposed any new framework to enhance this capability. In this work, we systematically study audio event relation modeling in TTA generation models. We first establish a benchmark for this task by: (1) proposing a comprehensive relation corpus covering all potential relations in real-world scenarios; (2) introducing a new audio event corpus encompassing commonly heard audios; and (3) proposing new evaluation metrics to assess audio event relation modeling from various perspectives. Furthermore, we propose a gated prompt tuning strategy that improves existing TTA models' relation modeling capability with negligible extra parameters. Specifically, we introduce learnable relation and event prompt that append to the text prompt before feeding to existing TTA models~\footnote{Code: https://github.com/yuhanghe01/RiTTA}. 
\end{abstract}

\vspace{-2mm}
\section{Introduction}
\vspace{-1mm}
\label{sec:intro}

Text-based crossmodal content generation has gained significant attention in recent years as it opens up new possibilities for even amateur users to create professional content. Typical such methods include text-to-image~(TTI)~\cite{ho2020denoising}, text-to-music~(TTM)~\cite{musicGEN}, text-to-point~(TTP)~\cite{point_E}, text-to-speech~(TTS)~\cite{fast_speech} text-to-audio~(TTA)~\cite{audioldm2,makeanaudio}. Among all of them, text-to-audio~(TTA) generation stands out as a particularly promising area, enabling the synthesis of complex acoustic environments or soundscapes directly from textual descriptions. Recent advances have demonstrated impressive progress in generating high-quality, detail-rich audio described in input text prompt~\cite{audioldm2,liu2023audioldm,makeanaudio,makeanaudio_v2,ghosal2023tango,ghosal2023tango2,kreuk2022audiogen}.

\begin{table}[t]
\small
\centering
\begin{tabular}{l|cl}
\hline
\multicolumn{3}{c}{\textit{\makecell[c]{\textbf{Text Prompt}: generate dog barking audio, \\followed by cat meowing audio}}}\\
\hline
Method & Rel? & Remark \\
\hline
AudioLDM~(\citeyear{liu2023audioldm}) & \xmark & \makecell[l]{just cat meow}\\
AudioLDM 2~(\citeyear{audioldm2}) & \xmark & output dog barking\\
MakeAnAudio~(\citeyear{makeanaudio}) & \xmark & just cat meow \\
AudioGen~(\citeyear{kreuk2022audiogen}) & \xmark & output wrong audios \\
Tango~(\citeyear{ghosal2023tango2}) & \xmark & \makecell[l]{two audios}\\
Tango~2~(\citeyear{ghosal2023tango2}) & \xmark & \makecell[l]{can output two audios}\\
TangoFlux~(\citeyear{tangoflux}) & \xmark & \makecell[l]{can not satisfy relation}\\
\hline
\end{tabular}
\vspace{-2mm}
\caption{A case study on existing TTA methods. ``Rel'' means ``if the relation is correctly modeled?''.}
\label{tab:relation_test}
\vspace{-7mm}
\end{table} 

When perceiving the physical world acoustically, whether through text or audio, the fundamental unit is the audio event, a distinct acoustic signal representing an independent source. The essence of perception lies in understanding the relationships emerging from events. Audio events are spatiotemporally distributed in the physical world. Together with relation, they contribute for holistic acoustic scene understanding~\cite{acoustic_scene_classify}. Studies in psychology~\cite{Zacks_Speer_Swallow_Braver_Reynolds_2007} and neuroscience~\cite{human_level_concept,hirsh1967brain} show that the human brain perceives the environment through discrete events and the relations between them. Humans are adept at using rich language to describe both audio events and their intricate relationships. While current TTA models can generate audios with high fidelity, their ability to generate audios that not only includes audio events but also preserves the text-informed relationships between them remains unexplored.

\begin{figure*}[t]
    \centering
    \includegraphics[width=0.98\linewidth]{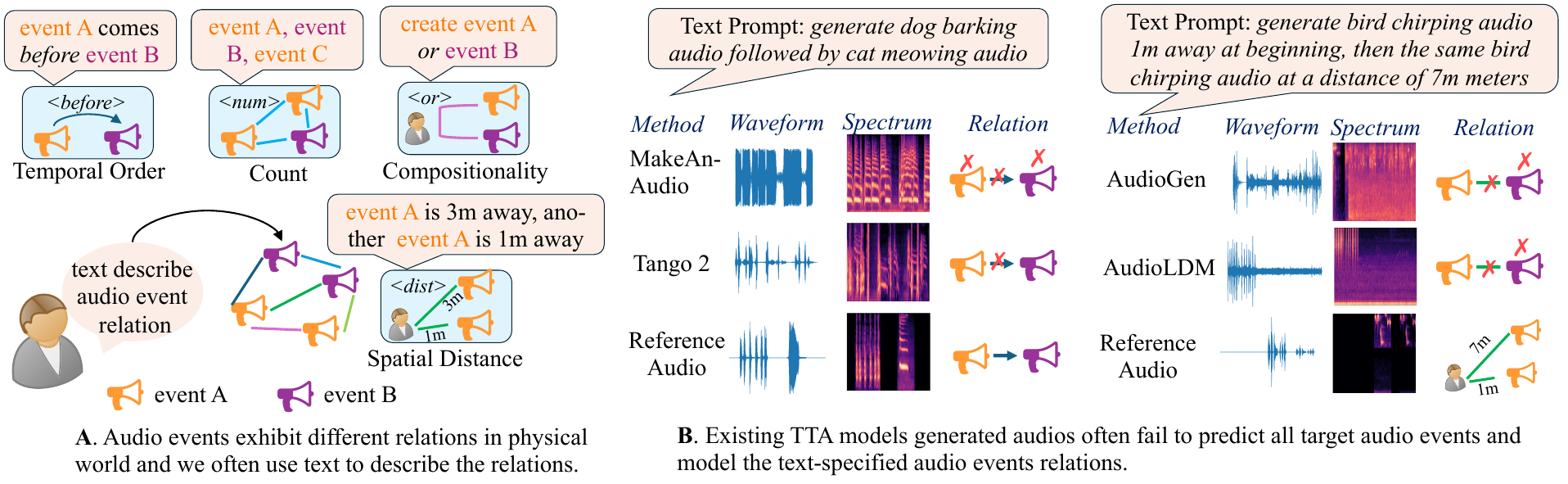}
    \vspace{-2mm}
    \caption{\emph{RiTTA} Motivation: The acoustic world is rich with diverse audio events that exhibit various relationships. While text can precisely describe these relationships~(Fig.~A), current TTA models struggle to capture both the audio events and the relations conveyed by the text~(Fig.~B). This challenge motivates us to systematically study \emph{RiTTA}.}
    \label{fig:teasing_fig}
    \vspace{-4mm}
\end{figure*}

As a primary study, we prompt the latest six TTA models with an exemplar text with explicit audio events and their relation \textit{generate dog barking audio, followed by cat meowing audio}. Next we check if the specified audio events are present and if so, their relations are correct in the generated audios. As is shown in Table~\ref{tab:relation_test}, all existing TTA models fail to properly model temporal relationships in the generated audio, even when they succeed in generating the correct audio events. The generated audio waveform, spectrum and another case study with a much complex text are shown in Fig.~\ref{fig:teasing_fig}. The poor performance of current TTA models in modeling audio events relation, along with the lack of systematic discussion on this topic, motivates us to explore \emph{Relation in TTA}~(dubbed \emph{RiTTA}) in depth in this work. We visualize the motivation in Fig.~\ref{fig:teasing_fig}.

To systematically study \emph{RiTTA}, we first benchmark it from four key perspectives: 1. we construct a comprehensive audio event relation corpus that captures common relationships found in the physical world. Unlike visual relations in cross-modal image tasks, which mainly focus on spatial aspects~(\textit{e.g.}, left, bottom)~\cite{gokhale2022benchmarking}, audio events exhibit far more complex relationships spanning spatial, temporal, and compositional dimensions. Consequently, we define four primary relation categories: \emph{Temporal Order}, \emph{Spatial Distance}, \emph{Count}, and \emph{Compositionality}. 2. Accompanying the relation corpus, we build an audio event category corpus derived from five main sources, each of which is further linked to multiple seed audios. 3. devise a \texttt{<text,audio>} pair generation strategy emphasizing both text prompt and audio diversity. 4. propose a new relation aware evaluation protocol that assesses the relation in a multi-stage manner. The proposed benchmark will benefit the community to explore \emph{RiTTA} in greater depth. Additionally, we introduce gated prompt tuning strategy to significantly improve existing TTA models' relation modeling capability by simply introducing a negligible parameters. 

\vspace{-3mm}
\begin{enumerate}[leftmargin=*]
    \item We conduct extensive evaluation on existing TTA models' inability in relation modeling.\vspace{-2mm}
    \item We benchmark \emph{RiTTA} by introducing three corpora: relation corpus, audio event corpus and seed audio corpus, as well as a new \texttt{<text,audio>} pair generation strategy.\vspace{-2mm}
    \item We propose a new multi-stage relation aware evaluation framework.\vspace{-2mm}
    \item We introduce gated prompt tuning to improve existing TTA models' relation modeling capability by introducing tunable prompts. \vspace{-2mm}
\end{enumerate}
\vspace{-2mm}

\section{Related Work}
\vspace{-1mm}

\textbf{Text-to-Audio~(TTA) Generation} involves producing audio that faithfully reflects the acoustic content or behavior described by the input text. Recent advancements have significantly improved the quality and intelligibility of generated audio~\citep{audioldm2,liu2023audioldm,kreuk2022audiogen,yang2022diffsound,ghosal2023tango,liao2024baton}. AudioLDM~\citep{liu2023audioldm} builds on latent space~\citep{stable_diffusion} to learn continuous representation. The most recent work TangoFlux~\cite{tangoflux} adopts flow matching to improve the performance. Despite the improvement, existing TTA methods still lag significantly in their ability to model relationships between audio events in the generated audio.

\textbf{Audio Events Relation Modeling} Based on how audio interact with the physical world in space, time and perceptual aspects, the resulting audio events exhibit complex relationships in spatial, temporal and compositional aspects. Prior work has partially addressed modeling certain temporal relations~(\textit{e.g.}, order) in TTA~\citep{audiotime} and compositional reasoning~\citep{ghosh2024compa} for discriminative tasks, such as audio classification and audio-text retrieval. While prior research has touched on modeling audio event relations, their potential in TTA remains largely underexplored. If we analogize an audio event to an object in image, the corresponding relationships exhibited in an image are mainly limited to 2D spatial relationship~(\textit{e.g.}, before, bottom, left). Despite object of interest spatial relationship learning and evaluation have received lots of attention in recent years~\citep{Krishna2016VisualGC,gokhale2022benchmarking,compos_ability}, the research on audio event relation modeling has been almost ignored.

\textbf{Prompt Tuning}~\cite{jia2022vpt,APT_prompttune,lester-etal-2021-power} is originally proposed in Natural Language Processing~(NLP)~\cite{lester-etal-2021-power} as an efficient alternative to full finetuning for large pre-trained models. Prompt tuning method proposes learnable prompts as task-specific continuous vectors that are directly optimized via gradients during fine-tuning. In recent years, prompt tuning has been successfully adopted in computer vision as visual prompt tuning~(VPT)~\cite{jia2022vpt,sohn2022visualprompttuninggenerative} and audio as audio prompt tuning~(APT)~\cite{APT_prompttune,apt_deepfake}. Inspired by the prompt tuning, we introduce gated prompt tuning strategy that significantly improves existing TTA models performance on relation-aware generation in a parameter-efficient way.
\vspace{-2mm}
\section{Benchmark Relation-Aware TTA}

\begin{figure*}[t]
\begin{minipage}[t]{.5\linewidth}
\centering
\small
    \renewcommand{\arraystretch}{0.5}
    \begin{tabular}{m{1.2cm}|m{1.4cm}|m{3.5cm}}
    \hline
    \makecell[c]{Main\\ Relation} & \makecell[c]{Sub-\\Relation} & \makecell[c]{Sample Text Prompt} \\
    \hline
       \makecell[c]{Temporal \\ Order}  &  \makecell[c]{before;\\ after;\\ simultaneity} & \makecell[c]{generate dog barking audio,\\ followed by cat meowing;}\\
       \hline
       \makecell[c]{Spatial \\Distance} & \makecell[c]{close first;\\far first;\\ equal dist.} & \makecell[c]{generate dog barking audio\\ that is 1 meter away, follow-\\ed by another 5 meters away.}\\
       \hline
      \makecell[c]{Count}& \makecell[c]{count} & \makecell[c]{produce 3 audios: dog bark-\\ing, cat meowing and talking.}\\
       \hline
       \makecell{Composit\\ionality} & \makecell[c]{and; or;\\ not;\\ if-then-else} & \makecell[c]{create dog barking audio\\ or cat meowing audio.}\\
       \hline
    \end{tabular}
    \vspace{-1mm}
    \captionof{table}{Audio Events Relation Corpus.}
    \label{tab:relation_corpus}
\end{minipage}\quad
\hspace{2mm}
\begin{minipage}[t]{.4\linewidth}
\centering
\small
    \begin{tabular}{m{1.7cm}|m{4.cm}}
    \hline
    \makecell[c]{Main\\Category} & \makecell[c]{Sub-Category} \\ 
    \hline
       \makecell[c]{Human\\ Audio}  &  baby crying; talking; laughing; coughing; whistling \\
       \hline
       \makecell[c]{Animal\\ Audio} & cat meowing; bird chirping; dog barking; rooster crowing; sheep bleating\\
       \hline
       \makecell[c]{Machinery} & boat horn; car horn; door bell; paper shredder; telephone ring\\
       \hline
       \makecell[c]{Human-Object\\ Interaction} &vegetable chopping; door slam; footstep; keyboard typing; toilet flush\\
       \hline
        \makecell[c]{Object-Object\\ Interaction} & emergent brake; glass drop; hammer nailing; key jingling; wood sawing\\
       \hline
    \end{tabular}
    \vspace{-1mm}
    \captionof{table}{Audio Events Category Corpus.}
    \label{tab:event_corpus}
\end{minipage}
\vspace{-4mm}
\end{figure*}

\subsection{Audio Event Relation Corpus}
\label{sec:relation_corpus}

An audio event refers to a distinct acoustic signal occurrence with specific frequency, duration and context characteristics that can be attributed to distinguish an independent sound source~\cite{sounddet} in an environment. Audio event is ubiquitous in the physical world and serves as the fundamental entity to analyze and interpret the acoustic scene. We embrace the audio event as the fundamental element to construct the relation corpus.

We construct the audio events relation corpus based on two key aspects. First, we consider relations commonly found in the physical world, such as those arising from spatial and temporal variations, which test TTA models' ability to replicate audio events' interactions in real-world scenarios. Second, we focus on relations that challenge TTA models' logical reasoning, evaluating their ability to determine both which audio events to generate and how to generate them. These two aspects partially overlap. Specifically, we define five main audio event categories, each associated with five subcategories of audio events. The detailed relation corpus is provided in Table~\ref{tab:relation_corpus}, including,

\begin{enumerate}[leftmargin=*]
    \item \textbf{Number Count}: The number of audio events included in audio, testing TTA models' ability to address acoustic polyphony challenge.\vspace{-2mm}
    \item \textbf{Temporal Order}: Temporal order refers to the sequence of audio events in the generated audio. We include three basic temporal relations for two audio events: \texttt{before}, \texttt{after}, and \texttt{simultaneity}, testing the TTA models' ability to distinguish and generate the correct event order as specified in the input text prompt.\vspace{-2mm}
    \item \textbf{Spatial Distance}: Spatial distance refers to the variation in relative spatial distances inferred from the generated audio. It evaluates the TTA models' ability to capture the spatial distance differences specified in the text prompt. Since we focus on mono-channel audio, obtaining the absolute distance for each audio event is nearly impossible~\cite{sounddet}. Therefore, we rely on loudness differences within intra-class audio events to verify their spatial distance variations.\vspace{-2mm}
    \item \textbf{Compositionality}: Compositionality relation describes how multiple individual audio events are integrated together to form a complex auditory structure that specified in the input text prompt. It tests TTA models' logical reasoning capability in determining which audio events to generate and how to structure them, by following the guidance illustrated in the input text prompt. Specifically, we incorporate four main compositionality relations: Conjunction~(\texttt{And}, \textit{e.g.}, generate audio A and audio B together); Disjunction~(\texttt{Or}, \textit{e.g.}, generate audio A or Audio B, not both); Negation~(\texttt{Not}, exclude one particular audio event, \textit{e.g.}, do not generate dog barking audio); Condition~(\texttt{if-then-else}, either generate two audio events if the condition is met, otherwise generate the third audio if the condition is not met). \vspace{-2mm}
\end{enumerate}

Most of the relations relate to two audio events~(see Table~\ref{tab:event_corpus} for more detail). Expanding to include more complex relations with a greater number of audio events is left for future work.

\subsection{Audio Event Category Corpus}
\label{sec:audio_event_category}

Alongside the relation corpus presented in Sec.~\ref{sec:relation_corpus}, we further construct a comprehensive audio event category corpus. The two corpora serve as fundamental dataset for constructing text prompts for TTA models. Since different audio event signals are generated from various sources or through different interactions, we first establish four main audio source categories, further detailing each category with five sub-categories. These constructed audio categories encompass the majority of ubiquitous audio events encountered in our daily lives. Specifically, the audio event category corpus contain,

\vspace{-2mm}
\begin{enumerate}[leftmargin=*]
    \item \textbf{Human Audio}: the audio generated by human beings in our daily life, including \textit{baby crying}, \textit{coughing}, \textit{laughing}, \textit{whistling}, \textit{female speech} and \textit{male speech}.\vspace{-2mm}
    \item \textbf{Animal Audio}: the audio generated by animals, including \textit{cat meowing}, \textit{dog barking}, \textit{bird chirping}, \textit{horse neighing}, \textit{rooster crowing}, \textit{sheep bleating} and \textit{pig oinking}.\vspace{-2mm}
    \item \textbf{Machinery Audio}: audio generated by various machinery devices while they are working, including \textit{car horn}, \textit{doorbell}, \textit{telephone ring}, \textit{paper shredder} and \textit{boat horn}.\vspace{-2mm}
    \item \textbf{Human-Object Interaction Audio}: human-object interaction audios include \textit{vegetable chopping}, \textit{keyboard typing}, \textit{toilet flushing}, \textit{door slamming} and \textit{foot step}.\vspace{-2mm}
    \item \textbf{Object-Object Interaction Audio}: we further incorporate object-object interaction audios, including \textit{glass dropping}, \textit{car emergency brake}, \textit{hammering nail}, \textit{wood sawing} and \textit{keys jingling}.\vspace{-2mm}
\end{enumerate}
The detailed audio event corpus is given in Table~\ref{tab:event_corpus}. With the constructed relation and audio event corpus, we can create relation aware text prompts for TTA models.

\section{Seed Audio and Text-Audio Pair Creation}
\label{sec:seed_audio_corpus}

\begin{figure}[h]
\begin{mdframed}[style=prompt]
\small
1.~generate audio A succeeded by B;\\
2.~start with A, followed by B;\\
3.~play A initially, B afterwards;\\
4.~generate A preceded by B;\\
5.~A in the beginning, B coming next;
\end{mdframed}
\vspace{-2mm}
\caption{GPT-4 augmented prompts~(\texttt{before} relation).}
\label{gpt-4-text}
\vspace{-3mm}
\end{figure}

In order to create the corresponding audio for any constructed text prompt, we instantiate each audio event presented in Sec.~\ref{sec:audio_event_category} in the main paper with five exemplar seed audios collected from \texttt{freesound.org}~\footnote{since \texttt{freesound.org} does not contain meaningful people talking audio, we collect people talking audio from VCTK~\cite{yamagishi2019vctk}}. Since most audio files on \texttt{freesound.org} are uploaded by volunteers who recorded them in their daily lives, incorporating five exemplar audios for each individual audio event category enhances both the diversity and realism of the seed audio. For instance, in the case of the \texttt{dog barking} audio event, the five selected audios vary in terms of dog breeds and barking styles. To further enhance an audio event's temporal length diversity, we randomly slice each seed audio into non-overlapping clips ranging from 1~sec to 5~secs. In summary, we have constructed 11 relations~(see Table~\ref{tab:relation_corpus} Sub-Relation column), and 25 audio events across five main audio events categories. Each audio event has been associated with 5 diverse audio clips ranging from 1~sec to 5~secs collected from \texttt{freesound.org} dataset.

\textbf{Text Prompt Generation}: a proper audio events relation aware text prompt comprises of two parts: a relation~(\textit{e.g.}, \texttt{<before>}) and audio events categories. The audio event categories can be either intra-class or inter-class, and the audio event number depends on the relation. We first instantiate an initial text prompt describing this relation. For example, for the temporal order \texttt{before} relation, the initial text prompt can be like: \textit{generate audio A, followed by audio B}. To enrich the text prompts, we further use the initial text prompt to query LLM~(in our case GPT-4) to provide more text prompts with diverse descriptive language for the same relation. 
One such GPT-4 augmented text prompts is shown in Fig.~\ref{gpt-4-text}, which illustrates that the same relation can be exactly expressed by multiple different text prompts. By incorporating GPT-4, we create 5 text prompts for each individual relation.

\begin{figure}[t]
    \centering
    \includegraphics[width=0.95\linewidth]{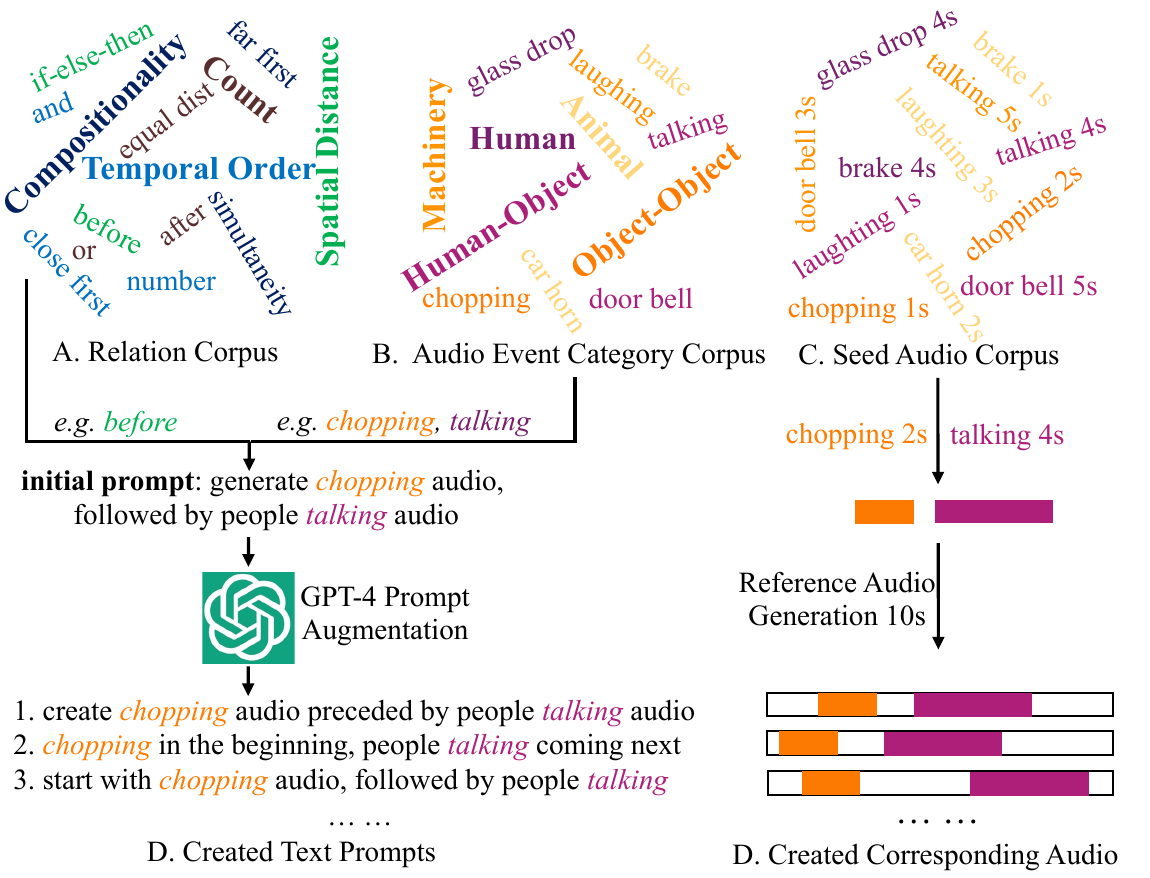}
    \caption{Relation aware \texttt{<textprompt,audio>} pair creation pipeline. It introduces large diversity in both text prompt and audio.}
    \label{fig:relation_datagen_pipeline}
\end{figure}

\textbf{Audio Generation}: Given the aforementioned audio events categories and relation, we randomly select an exemplar seed audio for each audio event and further linearly blend them together by satisfying the specified relation. For example, the relation \texttt{<before>} requires two audio events, the two selected audios can be blended together to form the final audio as long as the two seed audios satisfy the \texttt{<before>} relation~(Fig.~\ref{fig:relation_datagen_pipeline}, D). Notably, unlike blending two objects in an image that requires careful consideration of factors like occlusion and viewing angle, combining two audio signals simply involves linearly adding them together~\cite{wavetheory}. This offers an advantage for audio generation, as it eliminates the need for additional operations beyond the specified relation.

The generation of the \texttt{<text,audio>} pair is further illustrated in Fig.~\ref{fig:relation_datagen_pipeline}. With the proposed \texttt{<text,audio>} pair generation strategy, we can create massive diverse pairs even for the same audio events and the same relation, significantly enhancing the diversity and generalization capability of our generated dataset.

\subsection{Relation-Aware Evaluation Protocol}
\label{sec:relation_eval}

Existing TTA methods adopt general evaluation metrics to asses the similarity between generated audio and reference audio, including Fr\'{e}chet Audio Distance~(FAD), Fr\'{e}chet Distance (FD)~\cite{fid_distance}, Kullback–Leibler~(KL) divergence, Fr\'{e}chet Inception Distance~(FID) \textit{etc.}, among others. While those general evaluation metrics give an overall estimation of the similarity between the two comparing audios, they do not offer direct relation-aware evaluations. In addition to incorporating general evaluation metrics, we further propose multi-stage relation-aware evaluation metrics, with which we can gain insight on how the method performs w.r.t. difference relations.

\textbf{General Evaluation Metric}: We incorporate three widely used general evaluation metrics: the objective evaluation FAD, FD and KL divergence scores. 
FAD and FD measure the distribution similarity with feature embedding extracted from pre-trained on VGGish model~\cite{vggish}.

\begin{figure}
    \centering
    \includegraphics[width=0.90\linewidth]{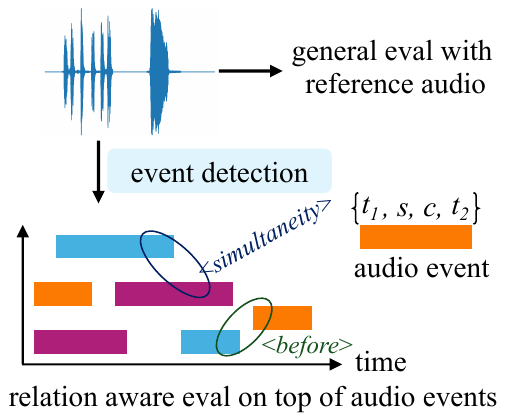}
    \vspace{-2mm}
    \caption{Relation aware evaluation. Audio event detection model is applied to get audio events. The meta data of each event contains start time $t_1$, end time $t_2$, confidence score $s$ and class label $c$. Various relations can be discovered from these audio events.}
    \label{fig:relation_eval_vis}
    \vspace{-3mm}
\end{figure}

\textbf{Relation aware Evaluation Metric}: To directly measure how accurately the text-indicated relation is reflected in the generated audio, we incorporate relation aware metrics for each specified relation.

In relation aware evaluation, we base on the individual audio event to compute the metrics, which allows us to measure the relation between audio events. Let's denote $(\mathcal{A}_g, \mathcal{T}, \mathcal{R}, \mathcal{A}_p)$ by ground truth audios, text prompts, relations and generated audios, respectively. We first extract audio events $\mathcal{E}$ from generated audios $\mathcal{A}_p$. For example, for the $i$-th generated audio $a_i^p$, we apply pre-trained audio event detection model~(we use finetuned PANNS~\cite{panns}, see Sec.~\ref{sec:panns_finetune} in Appendix) to extract all potential audio events involved in the audio $E_{a_i^p}=\{(e_j, m_j)|s\}_{i=1}^{k}$ by a given event confidence threshold $s\in \mathcal{S}$, where $e_j$ is the $j$-th audio event and $m_j$ is the corresponding meta data~(\textit{e.g.}, class label, confidence score, temporal start and end time, see Fig.~\ref{fig:relation_eval_vis}). To obtain audio events data for ground truth audios, we can either apply the same pre-trained model or directly extract from text prompts. Finally, we can get $(\mathcal{A}_g, \mathcal{T}, \mathcal{R}, \mathcal{A}_p, \mathcal{E}_p, \mathcal{E}_g)$, the relation aware evaluation function $f(\cdot)$ depends on the audio events $\mathcal{E}_p$, $\mathcal{E}_g$ and relations $\mathcal{R}$, $f(\mathcal{E}_p, \mathcal{E}_g|\mathcal{R}, s)$. We adopt a multi-stage relation aware evaluation strategy.

\textbf{Stage 1:} Target Audio Events Presence~(\textbf{Pre}). The paramount requirement for a successful audio generation is the presence of text-specified audio events in the generated audio. In this evaluation, the ground truth audio events and generated audio events are treated as \textit{set}. For a given ground truth and generated audio events pair $(E_g, E_p)$, we iterate over each audio event $e_g$ in the ground truth $E_g$ to check if it exists in the generated audio events $E_p$, regardless of its number and temporal position. 
\vspace{-4mm}
\begin{equation}
    \begin{aligned}
        f_p(E_p, E_g) &= \frac{1}{k} \sum_{e_g \in E_{g}} \mathds{1} (e_{g}, E_{p}); \\
        \mathds{1}(e_g, E_p) &= 
        \begin{cases}
            1, & \text{if } e_g \in E_p,  \\
            0, & \text{otherwise}.
        \end{cases}
    \end{aligned}
    \label{eqn:eval_presence}
\end{equation}

\noindent where $k$ is event number in the ground truth. $s_l(e_g)$ is a potential event meeting the confidence threshold in the generated audio. We select the event with the highest confidence score as the target.

\textbf{Stage 2:} Relation Correctness~(\textbf{Rel}). Once confirming the aforementioned target audio presence, we further investigate if these audio events obey text-specified relation. The relation is correctly modeled if at least a subset of generated audio events meet the relation. We give score 1 if relation is correctly modeled, otherwise score 0.

\begin{equation}
\begin{aligned}
    f_r(E_p|R) &= \max_{E_t \in E_p \cap E_g} \{\mathds{1}(E_t,R)\};\\
    \mathds{1}(E_t,R) &=
    \begin{cases}
        1,&\ \text{if}\ E_t\  \text{meets}\  R, \\
        0,& \ \ \text{otherwise},
    \end{cases}
\end{aligned}
\label{eqn:eval_relation}
\end{equation}

\textbf{Stage 3:} Audio Parsimony~(\textbf{Par}). Apart from requiring to generate all target audios, we should discourage the model from generating excessive intra-class audio events or irrelevant inter-class audio events. We call this property \textit{Audio Parsimony}. Once it is violated, we introduce extra penalty,

\begin{equation}
    f_s(E_p, E_g) = \exp{(-w_s\cdot|n(E_p) - n(E_g)|)}
\label{eqn:eval_parsimony}
\end{equation}

\noindent where $n(\cdot)$ indicates event number. $w_s$ is the weight adjusting the penalty~(in our case, $w_s=0.1$). The higher audio event number incurs lower parsimony score, the resulting parsimony score lies within $(0,1)$. The final relation aware score based on event confidence threshold $s$ equals to the multiplication of the three stage scores,

\vspace{-4mm}
\begin{equation}
    \begin{aligned}
        f(\mathcal{E}_p, \mathcal{E}_g|\mathcal{R}, s) &= \frac{1}{N} \sum_{(E_p, E_g, R) \in (\mathcal{E}_p, \mathcal{E}_g, \mathcal{R})} \\
        \quad f_p(E_p, E_g) & \cdot f_r(E_p|R) \cdot f_s(E_p, E_g)
    \end{aligned}
\end{equation}

\noindent where $N$ is data number. The final average MSR~(AMSR) score $f(\mathcal{E}_p, \mathcal{E}_g|\mathcal{R}, s)$ lies within $[0, 1)$~(the higher of the score, the better of the model's performance). Following prior COCO object detection evaluation strategy~\cite{coco_dataset}, we further average across multiple discrete audio event confidence thresholds to get the mean average MSR score~(mAMSR), $f(\mathcal{E}_p, \mathcal{E}_g|\mathcal{R})$,

\begin{equation}
    f(\mathcal{E}_p, \mathcal{E}_g|\mathcal{R}) = \frac{1}{K}  \sum_{s\in \mathcal{S}} f(\mathcal{E}_p, \mathcal{E}_g|\mathcal{R})
\label{eqn:mMSR}
\end{equation}
\noindent where $K$ is the discrete audio event confidence thresholds number. In our case we use uniformly sample four confidence thresholds in range $[0.5, 0.8]$ with step size $0.1$.

\section{Gated Prompt Tuning}

We introduce gated prompt tuning, a new strategy that enables parameter-efficient and task-adaptive relation-modeling without explicitly intervening existing TTA models' design. Specifically, drawing inspiration from recent advancement in prompt tuning~\cite{APT_prompttune,jia2022vpt}, we introduce learnable relation prompt for each relation and event prompt for each audio event class, and further feed these prompts alongside text prompts to the TTA neural network for optimization.

\begin{figure}[t]
    \centering
    \includegraphics[width=0.80\linewidth]{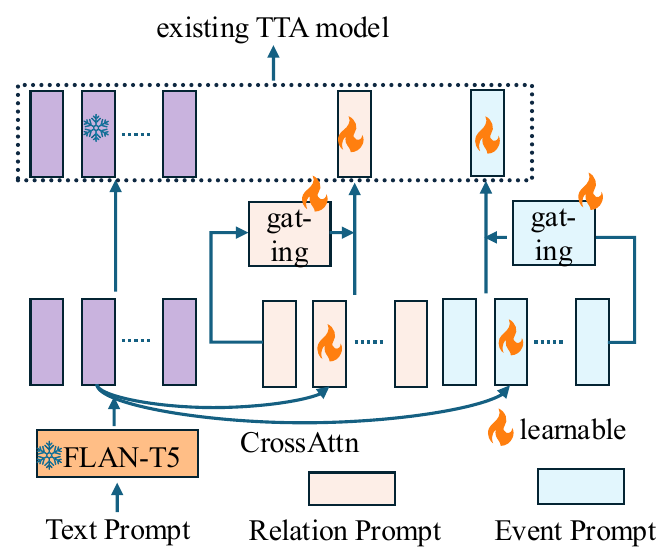}
    \vspace{-2mm}
    \caption{\small Gated Prompt Tuning Illustration.}
    \label{fig:ritta_gpt}
    \vspace{-3mm}
\end{figure}

Formally, we construct a learnable one-dimensional prompt for each relation and each audio event~$[P_r, P_e]$,($P_r\in \mathbb{R}^{T_r\times d}$, $P_e\in \mathbb{R}^{T_e\times d}$, $T_r$ is the relation numer, $T_e$ is the audio event class number), the prompt size equals to text prompt token embedding size~(in our case, 1024). Instead of directly concatenating all prompts with text prompt tokens, we first condition the learnable prompts on the input text tokens via cross attention~\cite{att_all_need}. Afterwards, we compute a gated combination for $P_r$ and $P_e$ separately, resulting in one integrated relation prompt and another integrated event prompt. By appending these two integrated prompts to text tokens and feeding them to an existing TTA model, we jointly tuning the prompts and the TTA model to instill relation modeling capability into the TTA model. 

As each input text just relies on sparse audio events and relations, we adopt the $\operatorname{entmax}_{1.5}$~\cite{entmax} gating mechanism to encourage the model to focus on a small subset of prompts. Unlike softmax, the $\operatorname{entmax}_{1.5}$ transformation yields sparse probability distributions, enabling some prompts to be assigned zero weights. To compute this gating, we first extract a summary representation of the prompts using mean average pooling, which is further fed to fully-connect layers to learn the gating logits~(the value before $\operatorname{entmax}_{1.5}$). With the $\operatorname{entmax}_{1.5}$ computed weight, we weight-sum the relation prompts~(or event prompts) to get one relation prompt~(or one event prompt, accordingly). The whole pipeline is illustrated in Fig.~\ref{fig:ritta_gpt}.

Since each added prompt is associated with a specific relation or event class, we add a classification loss to each learnable prompt during training phase to encourage each prompt to learn its designated class. It is worth noting that the introduced prompts' parameter size~(5~M) is negligible with comparing with the existing TTA model parameter size~(\textit{e.g.}, the Tango parameters is 866~M), and it does not intervene existing TTA model architecture. We experimentally show the effectiveness of this design in distilling relation modeling capability into existing TTA models.

\section{Experiment}

We run two experiments: benchmarking existing TTA methods on our curated 22~hrs dataset~(aka testing dataset); comparing gated prompt tuning strategy with other baselines to show its efficiency.

\begin{table*}[t]
\centering
\small
\begin{tabular}{c|l|p{1.0cm}<{\centering}|p{0.7cm}<{\centering}p{0.7cm}<{\centering}p{0.7cm}<{\centering}|p{0.7cm}<{\centering}p{0.7cm}<{\centering}p{0.7cm}<{\centering}|p{0.9cm}<{\centering}}
\hline
\multirow{2}{*}{\makecell[c]{Benchmark\\Method}} & \multirow{2}{*}{Model} & \multirow{2}{*}{\#param} & \multicolumn{3}{c|}{General Evaluation}  & \multicolumn{4}{c}{Relation Aware Evaluation~($\uparrow$)} \\
\cline{4-10}
 & &   & FAD~$\downarrow$ &  KL~$\downarrow$ & FD~$\downarrow$ & mAPre& mARel & mAPar &mAMSR \\
\hline
\multirow{8}{*}{Zero-Shot} & AudioLDM~(S-Full)~(\citeyear{liu2023audioldm})  & $185$~M  & \cellcolor{secondcolor}5.65  & 38.95 & \cellcolor{thirdcolor}37.30 & 2.76 & 0.50 & 2.52 & 0.04\\
&AudioLDM~(L-Full)~(\citeyear{liu2023audioldm})  & $739$~M  & \cellcolor{topcolor}5.47  & 38.42 & 37.96 & 3.09 & 0.77 & 2.56 & 0.08\\
&AudioLDM~2~(L-Full)~(\citeyear{audioldm2}) & $844$~M & 6.68 & \cellcolor{secondcolor}29.07 &  \cellcolor{secondcolor}35.85& 12.26& 2.41 & 10.01 & 3.39 \\
&MakeAnAudio~(\citeyear{makeanaudio}) & $452$~M & 9.46 & 82.72  & 45.98 & 8.14  & 1.68 & 6.47 & 1.02\\
&AudioGen~(\citeyear{kreuk2022audiogen}) & $1.5$~B & \cellcolor{thirdcolor}6.43 & \cellcolor{topcolor}28.01& \cellcolor{topcolor}32.04 & 9.61 & 2.12 & 8.60 & 2.27\\
&Tango~(\citeyear{ghosal2023tango}) & $866$~M & 10.79 & 90.26 & 39.46 & 11.13 & 2.27 & 9.88 & 3.10\\
&Tango~2~(\citeyear{ghosal2023tango2}) & $866$~M & 13.84  &  89.66 & 44.03 & 16.63 & 4.40 & 12.53 & 11.55 \\
&TangoFlux~(\citeyear{tangoflux}) & 515~M & 8.07 & \cellcolor{thirdcolor}32.80 & 47.92 & \cellcolor{secondcolor}33.83 & 7.02 & \cellcolor{topcolor}29.01 & \cellcolor{secondcolor}76.57 \\
\hline
\multirow{3}{*}{\makecell[c]{LLM+Agentic\\+TTA Model}} &Tango~(\citeyear{ghosal2023tango}) &  866~M & 11.88 & 92.19 & 41.44  & 12.33	&  \cellcolor{thirdcolor}9.21 & 11.28	& 19.17 \\ 
& Tango~2~(\citeyear{ghosal2023tango}) &  866~M & 14.76 & 93.10 & 44.89  & \cellcolor{thirdcolor}19.33 & \cellcolor{secondcolor}9.37 & \cellcolor{thirdcolor}14.13	& \cellcolor{thirdcolor}20.31 \\
& TangoFlux~(\citeyear{tangoflux}) & 515~M &  8.93 & 32.99 & 49.00  & \cellcolor{topcolor}35.19	& \cellcolor{topcolor}9.69 & \cellcolor{secondcolor}28.22 & \cellcolor{topcolor}{79.43} \\
\hline
\end{tabular}
\vspace{-1mm}
\caption{Benchmark quantitative result across all relations. mAPre, mARel and mAPar are in $10^{-2}$. mAPre and mARel can be treated as \textit{presence}, \textit{relation correctness} percentage ratio, in range $[0, 100]$. mAPar score also lies within $[0, 100]$. mAMSR~($10^{-4}$) lies in range~$[0,1]$. The LLM is deligated by GPT-4. The \colorbox{topcolor}{top-}, \colorbox{secondcolor}{second-} and \colorbox{thirdcolor}{third-} performing methods are labeled in different colors.}
\vspace{-2mm}
\label{tab:general-results}
\vspace{-1mm}
\end{table*}

\subsection{More Discussion on Data Creation}

\label{sec:data_prepare}

We follow the strategy presented in Sec.~\ref{sec:seed_audio_corpus} to create the dataset. Specifically, for each of the 11 sub-relations in Table~\ref{tab:relation_corpus}, we create 720~(2~hrs) \texttt{<text,audio>} pairs for testing~(aka benchmark dataset) and 1440 pairs~(4~hrs audio) for training~(aka finetuning dataset). The highlight of training/testing dataset is in Table~\ref{tab:ritta_datacreate} Appendix.

To ensure all relations can be effectively evaluated, we applied two key constraints during the data creation process. First, to make the audio events countable without ambiguity, we selected inter-category audio events to form the \texttt{<text,audio>} pairs. This avoids the ambiguity that arises when using intra-category events, especially for those with repetitive, similar local occurrences (\textit{e.g.}, multiple instances of dog barking). Second, for the \emph{Spatial Distance} relation, we introduced a temporal order constraint to ensure that the two audio events do not overlap in time. Temporal overlap would require complex source separation models~\cite{source_separation_icassp} to distinguish individual events. By enforcing this non-overlapping constraint, the evaluation of \emph{Spatial Distance} becomes manageable using an audio event detection model~(see Sec.~\ref{sec:panns_finetune} in Appendix). To make all the proposed relation measurable, we approximate spatial distance by loudness distance. More evaluation setup is given in Sec.~\ref{discuss_ritta_eval} in Appendix.

\subsection{Relation-Aware Benchmarking Result}

\begin{table*}[t]
\centering
\small
\begin{tabular}{l|p{0.5cm}<{\centering}p{0.5cm}<{\centering}p{0.5cm}<{\centering}|p{0.6cm}<{\centering}p{0.6cm}<{\centering}p{0.7cm}<{\centering}|p{0.9cm}<{\centering}|p{0.5cm}<{\centering}p{1.2cm}<{\centering}p{0.9cm}<{\centering}p{.7cm}<{\centering}}
\toprule
\multirow{2}{*}{Model}  & \multicolumn{3}{c|}{General Evaluation}  & \multicolumn{4}{c|}{Relation Aware Evaluation~($\uparrow$)} &  \multicolumn{4}{c}{mAMSR Across Four Main Relations}\\
\cline{2-12}
 & FAD~$\downarrow$ &  KL~$\downarrow$ & FD~$\downarrow$ & mAPre& mARel & mAPar &mAMSR &  \emph{Count} & \emph{TempOrder} & \emph{SpatDist} & \emph{Compos} \\
\hline
%Tango~(\citeyear{ghosal2023tango}) &  10.79 &  90.26 &  39.46 & 11.13 &  2.27 &  9.88 &  3.10 & 0.16 & 3.44 & 0.82 & 8.10\\
Tango~(finetune) & 4.60  &  23.92 & 27.03 & 21.23 & 10.78 & 20.35 & 48.67 & 8.04 & 324.10 & 1.88 & 44.42 \\
Tango~(ours GPT) & 3.12  &  20.21 & 25.11 & 25.77 & 15.38 & 27.19 & 59.13 & 10.11 & 378.90 & 3.12 & 54.87 \\
\midrule
TangoFlux~(finetune) & 2.94  &  20.10 & 21.09 & 37.12 & 11.11 & 33.99 & 83.44 & 13.56 & 368.77 & 5.10 & 58.88 \\
TangoFlux~(ours GPT) & \cellcolor{topcolor}{1.60}  & \cellcolor{topcolor}17.98 & \cellcolor{topcolor}18.20 & \cellcolor{topcolor}43.11 & \cellcolor{topcolor}15.33 & \cellcolor{topcolor}39.10 & \cellcolor{topcolor}127.98 & \cellcolor{topcolor}13.04 & \cellcolor{topcolor}425.98 &\cellcolor{topcolor} 3.10 & \cellcolor{topcolor}69.56 \\
\bottomrule
\end{tabular}
\vspace{-2mm}
\caption{The quantitative comparison across general and relation-aware evaluation between finetuing strategy and our proposed gated prompt tuning strategy. The gated prompt tuning~(GPT) just introduces extra 5~M parameters.}
\vspace{-2mm}
\label{tab:finetune_result}
\end{table*}

\begin{figure*}[t]
    \centering
    \includegraphics[width=0.98\linewidth]{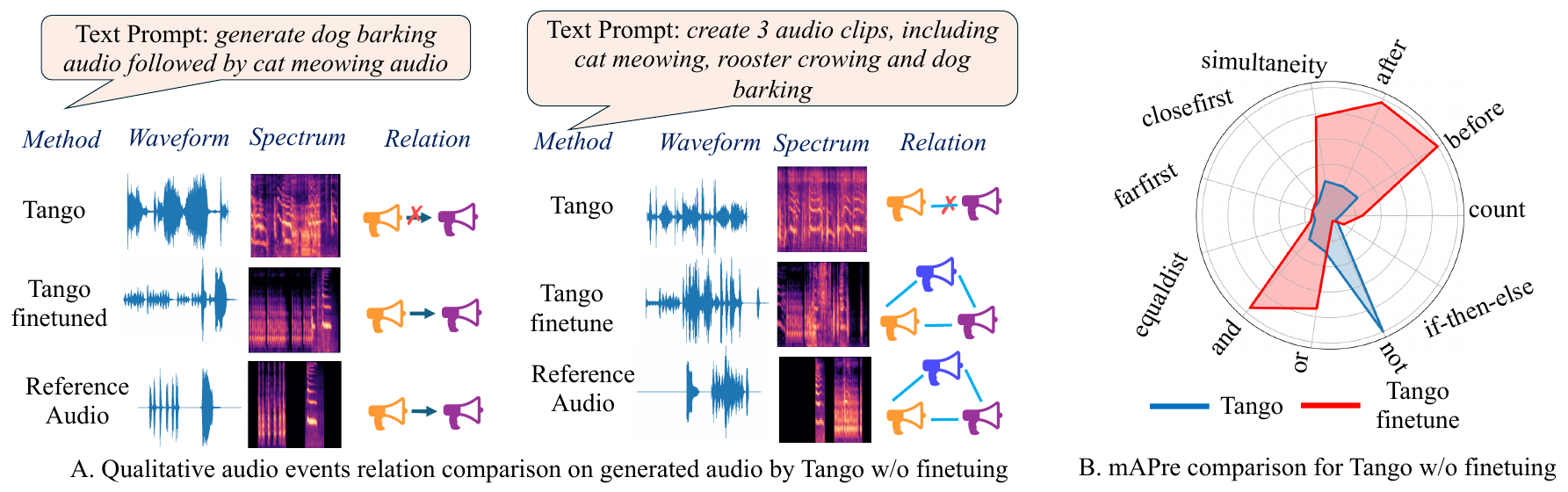}
    \vspace{-2mm}
    \caption{Qualitative visualization comparison of Tango w/o finetuning~(A) and mAPre w.r.t. 11 sub-relations.}
    \label{fig:tango_finetune_vis}
    \vspace{-3mm}
\end{figure*}

We benchmark our curated test dataset on 8 most recent TTA models: AudioLDM~\cite{liu2023audioldm}~(two versions), AudioLDM~2~\cite{audioldm2}, MakeAnAudio~\cite{makeanaudio}, AudioGen~\cite{kreuk2022audiogen}, Tango~\cite{ghosal2023tango}, Tango~2~\cite{ghosal2023tango2} and TangoFlux~\cite{tangoflux}. We directly depend on their released models to generate a 10~second audio from each text prompt, detailed configuration is given in Table~\ref{app:tab:model_setting} in Appendix. We further benchmark agentic workflow based methods, in which we leverage LLM~(GPT-4) acting as an agent to analyze the input text and output the separate audio events an TTA model needs to generate. At the same time, the same LLM works as the third agent to output the python code that merges the audios generated by the TTA model. The reason of experimenting agentic flow is to see if we can decompose the relation-aware generation task into simple single audio event generation task.

The quantitative evaluation results across all relations are shown in Table~\ref{tab:general-results}. From this table we can observe that the general evaluation results are inconsistent with our proposed relation aware evaluation metrics. The best performing methods under generational evaluations~(the two AudioLDM versions) perform the worst under relation aware evaluations, and vice versa. These discrepancies highlight the necessity of proposing evaluation metrics specifically tailored for audio events relations. Additionally, while the performance differences among the seven benchmarking methods under general evaluation are relatively minor, the corresponding differences under relation aware evaluation are significantly more pronounced~(\textit{e.g.}, Tango~2 outperforms AudioLDM~(S-Full) by about 200 times). However, even the top-performing method, TangoFlux~\cite{tangoflux}, still struggles to model audio events relations, as both its presence accuracy and relation accuracy rate are small~(less than $30\%$ accuracy rate on average). Moreover, agentic workflow performs slightly better under relation-aware evaluation but worse under general evaluation, which shows LLM involved agentic workflow cannot address relation modeling sufficiently.  All of these observations demonstrate the limitations of existing TTA methods in modeling audio events relation and the necessity to systematically study audio events relation in TTA, highlighting the importance of our proposed work. More experimental result is given in Sec.~\ref{key_findings} in Appendix.

In summary, we conclude that, 1. existing TTA models lack the ability to model audio events relation described by the text prompt in the generated audio, emphasizing the importance of our work in systematically study audio events relation in TTA. 2. Existing TTA evaluation metrics fall short in accurately measuring audio events relations from the generated audio. Our proposed multi-stage relation evaluation framework suffices to measure the relation accuracy from various aspects. 3. LLM involved agentic workflow does not suffice to address relation modeling.

\vspace{-2mm}
\subsection{Gated Prompt Tuning Result}

We run gated prompt tuning on two most advanced TTA models, Tango and TangoFlux, by initializing their model weights from pretrained model. All the learnable prompts are randomly initialized, each prompt is of size 1024. For Tango, we tune learnable prompts~(5~M) and latent diffusion model~(UNet, 866~M). For TangoFlux, we tune learnable prompts and Transformer blocks~(515~M). We use Adam optimizer with the learning rate $3\times 10^{-5}$, batch size of 16, SNR gamma value 5. We finetune 40 epochs on 4 A100 GPUs. The results, shown in Table~\ref{tab:finetune_result}, show that 1. finetuning either Tango or TangoFlux results in significant performance improvement in both relation-aware and general evaluation, and TangoFlux achieves better performance than Tango. 2. The introduced gated prompt tuning further improves the performance drastically. Considering the fact that gated prompt tuning just introduced negligible parameters, the performance gain directly shows the effectiveness of our proposed gated prompt tuning design.

Two qualitative examples of w/o finetuning Tango are in Fig.~\ref{fig:tango_finetune_vis}~A. It is evident that the finetuned Tango successfully models the \texttt{<before>} relation~(Table~\ref{tab:relation_test} and Fig.~\ref{fig:teasing_fig} show all existing TTA models fail on this case), and \texttt{<count>} relation. The mAPre score w.r.t. the 11 600
sub-relations is shown in Fig.~\ref{fig:tango_finetune_vis}~B (the mARel, mA- 601
Par, mAMSR are in Fig.~\ref{fig:tango_finetune_vis_append} in Appendix).

\subsection{Ablations on Gated Prompt Tuning}

\begin{table}[h]
    \centering
    \small
    \begin{tabular}{p{1.cm}<{\centering}p{1.cm}<{\centering}p{1.cm}<{\centering}p{1.cm}<{\centering}p{1.2cm}<{\centering}}
    \toprule
        GPT-only & GPT-Rel & GPT-Event & PT-only & GPT~(ours) \\
        \hline
        40.12 & 91.98 & 87.65 & 102.33 & \cellcolor{topcolor}{127.98} \\ 
        \bottomrule
    \end{tabular}
    \vspace{-1mm}
    \caption{Ablation Study Result on TangoFlux Model. We report mAMSR~($\uparrow$) score~($10^{-4}$).}
    \label{tab:ablation_study}
    \vspace{-4mm}
\end{table}

We run 4 main ablations on TangoFlux~\cite{tangoflux} based gated prompt tuning~(GPT) to validate the effectiveness of our gated prompt tuning design. 1. \textbf{GPT-only}: just train gated prompts without tuning the existing TTA model, which helps to test if tuning existing TTA model is necessary; 2. \textbf{GPT-Rel}, just introduce learnable relation prompts, which helps to test if jointly tuning event prompts is necessary; 3. \textbf{GPT-Event}, just introduce learnable event prompts, which in turn tests if jointly tuning relation prompts is necessary; 4. \textbf{PT-only}, prompt tuning without gating mechanism~(just mean-average all prompt to obtain the final single prompt). 

The mAMSR metric is given in Table~\ref{tab:ablation_study}. From this table we can observe that just tuning gated prompts leads to obvious performance drop~(it performs even worse than zero-shot based benchmark, see Table~\ref{tab:general-results}). We assume that such large difference results from the domain gap between conventional TTA and our introduced relation-aware TTA task, which naturally requires to tune large number of parameters to fill in the gap; Discarding either relation prompts~(GPT-Event) or events prompts~(GPT-Rel) also leads to obvious performance drop; Removing gating mechanism also sees performance drop~(but still performs better than GPT-Rel and GPT-Event); All of these ablation study results shows the importance of each of our introduced gated prompt tuning component.

\vspace{-1mm}
\section{Conclusion}
\vspace{-1mm}

We demonstrate existing TTA models struggle with relation modeling. Despite its importance, relation modeling has received rare attention in previous research. To address this gap, we introduce a new benchmark, a relation-aware evaluation metric, and a gated prompt tuning strategy. More discussion is in Sec.~\ref{sec:appen:conclusion} in Appendix.

\section{Limitations}

There are two main limitations in this work.

First, in this work, we incorporate 11 relation and 25 audio event in the relation corpus and event corpus, respectively. They are not sufficient enough to reflect the potential relations and audio events existing in the real scenarios. It is desirable to scale up the benchmark by introducing more audio event categories and accommodating more complex relations~(\textit{e.g.}, the nested combination of the 25 relations to generate more complex relation).

Second, a powerful relation-aware TTA model should be scalable to extend to incorporate new relations or new audio events automatically~(openended setting). Our current setting is close-ended, disallowing novel audio events or relation handling. This openended relation-aware TTA model also remains as a future research direction.

\bibliography{custom}
\clearpage

\appendix
\section*{Appendix}
\renewcommand{\thetable}{\Roman{table}}
\renewcommand{\thefigure}{\Roman{figure}}
\renewcommand\thesection{\Alph {section}}

\setcounter{section}{0}
\setcounter{figure}{0}
\setcounter{table}{0}

% \section{Finetuning PANNS Audio Event Detection Model on Our Curated Dataset}
% \label{sec:app:panns_finetune}

\section{Audio Event Detection Model Finetune}
\label{sec:panns_finetune}

\begin{figure}[h]
    \centering
    \includegraphics[width=0.99\linewidth]{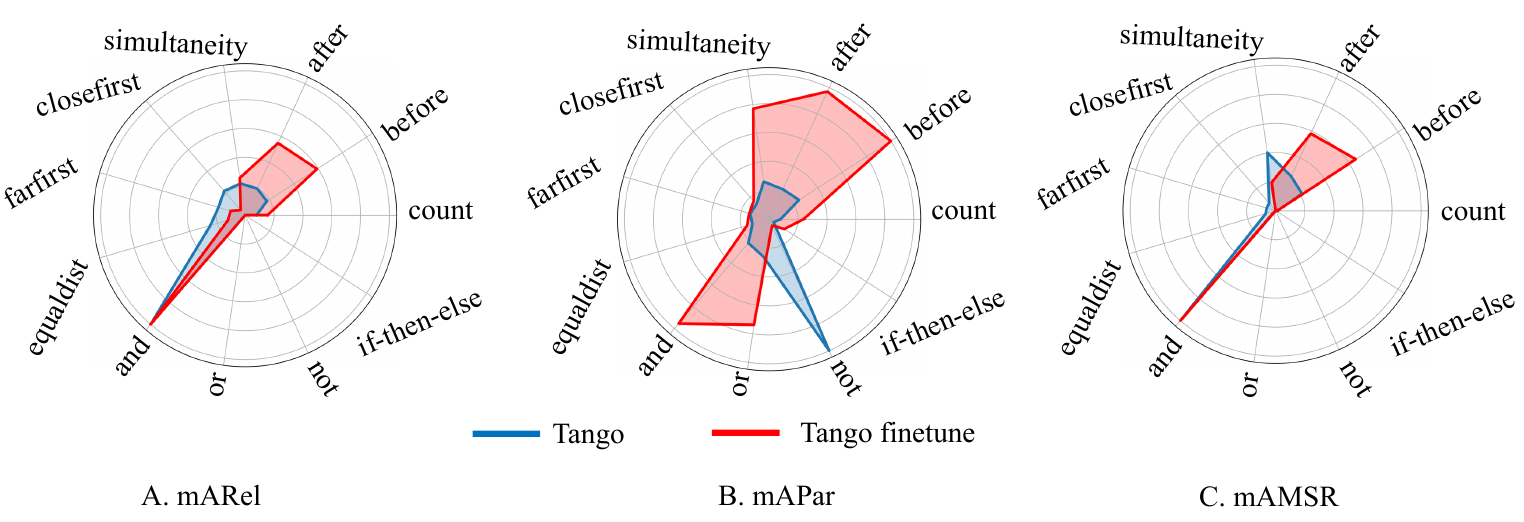}
    \caption{The comparison of mARel, mAPar, mAMSR on Tango w/o finetuning.}
    \label{fig:tango_finetune_vis_append}
\end{figure}

To detect the audio events from generated audio, we employ a pre-trained audio event detection model~(in our case, we adopt PANNS~\citep{panns}) to detect all audio events, each detected event has class label with a confidence score, start time and end time. Analyzing these detected audio events can uncover various audio events relations~(see Fig.~\ref{fig:relation_eval_vis} in the main paper).

\begin{table*}[t]
    \centering
    \begin{tabular}{l|c}
    \hline
    Methods & Setting \\
    \hline
    AudioLDM~(S-Full)~(\citeyear{liu2023audioldm})  & guidance\_scale=5, random\_seed=42, n\_candidates=3\\
AudioLDM~(L-Full)~(\citeyear{liu2023audioldm})  & guidance\_scale=5, random\_seed=42, n\_candidates=3 \\
AudioLDM~2~(L-Full)~(\citeyear{makeanaudio}) & guidance\_scale=3.5, random\_seed=45, n\_candidates=3\\
MakeAnAudio~(\citeyear{makeanaudio}) & ddim\_steps = 100, scale = 3.0\\
AudioGen~(\citeyear{kreuk2022audiogen}) & model name: audiogen-medium\\
Tango~(\citeyear{ghosal2023tango}) & num\_steps = 200, guidance=3, num\_samples=1\\
Tango~2~(\citeyear{ghosal2023tango2}) & num\_steps = 200, guidance=3, num\_samples=1\\
\hline
    \end{tabular}
    \caption{Detail setting for each TTA method.}
    \label{app:tab:model_setting}
\end{table*}

The PANNS model~\citep{panns} is pre-trained on the large-scale 527 class AudioSet dataset~\citep{audioset_dataset}. It contains an audio tagging model and an audio event detection model. Directly applying the pre-trained detection model to detect audio events from our generated audios inevitably results in false positive and ambiguous detections. For instance, a \textit{door slam} sound may be incorrectly detected as speech or music with high confidence scores. To mitigate the ambiguity and inaccuracies, we finetune the detection model~(``Cnn14\_DecisionLevelMax'' variant) on our specially curated 100~k dataset by just tuning the last classification layer. Finally the finetuned model achieves mAP~$0.57$ on our curated 10k test sets, far outperforming the original model with mAP~$0.43$.

We based on the pretrained PANNS~\citep{panns} audio event detection model to finetune it on our curated 100~k audio training dataset. Each audio is 10~s long with sampling rate 16~kHz. Moreover, each audio randomly contains one to five audio events, each event has a random start time position in the 10~s long audio. The input is 10~s long audio waveform. The output is a confidence map of shape $[20, 25]$, where 20 is the time steps with the temporal resolution 0.5~s and 25 is the audio event class number. Potential audio events are extracted from the confidence map by thresholding the confidence map, audio events with too short time duration~(in our case, less than 0.5~s) are discarded. The training and testing datasets size are 100~k and 10~k respectively. We adopt Adam~\citep{adam_optimizer} to train the model with initial learning rate 0.0001 but decays every 200 epochs with decaying rate 0.5. Finally, we train 350~epochs. The loss function is binary cross-entropy loss~(BCE). On the testing dataset, the finetuned model achieves mAP 0.57. We use the finetuned audio event detection model to detection audio events from the generated audios.

\section{More Discussion on RiTTA Evaluation}
\label{discuss_ritta_eval}

We specifically adjust the audio generation process for relations under \emph{Compositionality} and \emph{Spatial Distance} to so as to ensure these relations can be accurately evaluated under our proposed framework.

First, we skip general evaluation for \texttt{<Not>} as it lacks a corresponding ground truth reference audio. During fintuning, we generate silent audio for \texttt{<Not>} for create finetuing pairs. Second, for the \texttt{<if-then-else>} and \texttt{<Or>} sub-relations, which correspond to two possible ground truth audios, we handle evaluation by computing the L2 distance~(in the time domain) between the generated audio and the two reference audios. For example, for the prompt \textit{if event A then event B, else event C}, the first reference is the combination of events A and B, while the second contains only event C. We use the reference audio with smaller L2 distance to the generated audio for general evaluation.

Third, precise evaluation of the three sub-relations~(\texttt{<closefirst>}, \texttt{<farfirst>}, and \texttt{<equaldist>}) under \emph{Spatial Distance} from unconstrained audio requires sound event detection and localization~(SELD~\cite{soundsynp,seld_dcase19}) techniques to spatially localize each audio event, which is impossible with mono-channel audio. To address this, we approximate spatial distance by calculating the loudness, which can be estimated using the L2 norm of the audio waveform. The rationale behind this approach is that greater distances result in a dampening of waveform amplitude~(and consequently reduced loudness) due to energy decay along the audio propagation path. When the loudness difference exceeds a predefined threshold~(for \texttt{<closefirst>}, \texttt{<farfirst>}) or is within that threshold~(for \texttt{<equaldist>}), we consider the evaluation accurate. Specifically, we use a loudness reduction ratio $\sigma_1$ (with $\sigma_1=0.2$ in our case). For \texttt{<closefirst>}, if the closer event's loudness is at least $\sigma$ times greater than the further event's loudness, the relation is considered correct. Similarly, for \texttt{<equaldist>}, the loudness difference between the two events should be within $\sigma_2$~(with $\sigma_2=0.4$ in our case) of the louder event's loudness. This estimation is also reflected in the data generation process (see Sec~\ref{sec:data_prepare}).

\section{Existing TTA model Setting}
\label{sec:app:tta_setting}

We test 8 most recent TTA models: AudioLDM~\citep{liu2023audioldm}~(two versions), AudioLDM~2~\citep{audioldm2}, MakeAnAudio~\citep{makeanaudio}, AudioGen~\citep{kreuk2022audiogen}, Tango~\citep{ghosal2023tango}, Tango~2~\citep{ghosal2023tango2} and TangoFlux~\cite{tangoflux}. We depend on their released pre-trained model and use their recommended hyperparameter setting for benchmarking~(from their Github page). The detailed setting for each TTA method is given in Table~\ref{app:tab:model_setting}.

\section{More Result on Tango Finetuning}

The mARel, mAPar and mAMSR score w.r.t. 11 sub-relations is given in Fig.~\ref{fig:tango_finetune_vis_append}.

\begin{table*}[t]
\centering
\small
\caption{Benchmark quantitative result w.r.t. the four main relations. We report FAD sore and mAMSR score for general evaluation and relation aware evaluation, respectively.}
\begin{tabular}{l|p{0.5cm}<{\centering}p{1.1cm}<{\centering}p{.8cm}<{\centering}p{.8cm}<{\centering}|p{0.5cm}<{\centering}p{1.1cm}<{\centering}p{0.8cm}<{\centering}p{.8cm}<{\centering}}
\hline
\multirow{2}{*}{Model} & \multicolumn{4}{c|}{General Evaluation~(FAD $\downarrow$)}  & \multicolumn{4}{c}{Relation Aware Eval.~(mAMSR $\uparrow$)} \\
\cline{2-9}
 &  \emph{Count} &  \emph{TempOrder} & \emph{SpatDist} & \emph{Compos} &  \emph{Count} & \emph{TempOrder} & \emph{SpatDist} & \emph{Compos}  \\
\hline
AudioLDM~(S-Full)~\cite{liu2023audioldm}   & \cellcolor{secondcolor}{3.85} & \cellcolor{secondcolor}6.86 &  \cellcolor{thirdcolor}4.56 & \cellcolor{secondcolor}9.36 & 0.00 & 0.05 & 0.00 & 0.18 \\
AudioLDM~(L-Full)~\cite{liu2023audioldm}  & \cellcolor{topcolor}3.68  & \cellcolor{topcolor}6.45  & \cellcolor{secondcolor}4.10 & \cellcolor{topcolor}8.98 & 0.00 & 0.05 & 0.06 & 0.17\\
AudioLDM~2~(L-Full)~\cite{makeanaudio} & \cellcolor{thirdcolor}5.03 & 8.94 &  4.72 & \cellcolor{thirdcolor}9.41 & 0.14 & 1.87 & \cellcolor{secondcolor}1.46 & \cellcolor{secondcolor}9.89\\
MakeAnAudio~\cite{makeanaudio} & 6.02 & 10.21 &  8.18 & 12.78 & 0.12 & 0.66 & 0.44 & 2.40 \\
AudioGen~\cite{kreuk2022audiogen}& 6.14 & \cellcolor{secondcolor}8.39 & \cellcolor{topcolor}3.38 & 9.98 & \cellcolor{secondcolor}0.32 & \cellcolor{secondcolor}3.83 &  0.48 & 4.18\\
Tango~\cite{ghosal2023tango} & 8.54 &  10.25 & 10.11 & 13.97 & \cellcolor{thirdcolor}0.16 & \cellcolor{thirdcolor}3.44 & \cellcolor{thirdcolor}0.82 & \cellcolor{thirdcolor}8.10 \\
Tango~2~\cite{ghosal2023tango2} & 10.01  & 13.91 & 13.23 & 17.04 & \cellcolor{topcolor}0.96 & \cellcolor{topcolor}20.92 & \cellcolor{topcolor}1.92& \cellcolor{topcolor}23.25\\
TangoFlux~\cite{tangoflux} & 6.02  & 8.00 & 52.20 & 60.22 & 1.72 & 111.51 & 0.00& 47.71\\
\hline
\end{tabular}
\label{tab:results-fourcates}
\end{table*}

\begin{figure*}[t]
    \centering
    \includegraphics[width=0.98\linewidth]{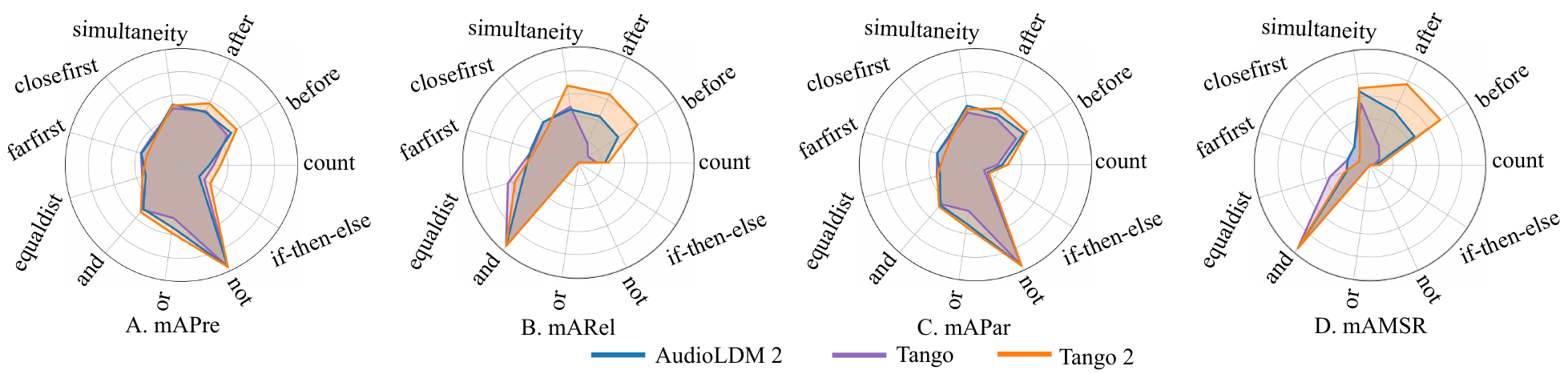}
    \caption{\small Top~3 performing in audio events relation modelling TTA methods' performance w.r.t. the 11 sub-relations. We report mAPre, mARel, mAPar and mAMSR scores separately.}
    \label{fig:exp_radar_chart}
\end{figure*}.

\section{\texttt{<\text{Text,Audio}>} pair generation and RiTTA benchmark highlight}

The RiTTA \texttt{<Text,Audio>} pair generation pipeline is illustrated in Fig.~\ref{fig:relation_datagen_pipeline} and RiTTA benchmark summary is highlighted in Table~\ref{tab:ritta_datacreate}.

\begin{table}[h]
    \centering
        \begin{tabular}{p{2.0cm}<{\centering}|p{5cm}<{\centering}}
        \hline
        Entry & Highlight\\
        \hline
        seed audio & \makecell[c]{one event has 5 audios\\
                                  each has 1~s-5~s audio clips}\\
        \hline
        \makecell[c]{audio categ-\\ory corpus} & \makecell[c]{5 main categories\\25 sub-categories}\\
        \hline
        \makecell[c]{relation\\corpus} & 4 main 11 sub relations\\
        \hline
        \makecell[c]{relation -\\event number} & \makecell[c]{\texttt{count}: 2-5 events;\\
                                                               \texttt{Not}: 1 event;\\
                                                               \texttt{if-then-else}: 3 events\\
                                                               others: 2 events.}\\
                                                               
        \hline
        \makecell[c]{train and \\ test data info} & \makecell[c]{each audio is 10~s long\\sampling rate 16~kHz\\
                                                                 train: 44~hrs, 1.6~k pairs\\
                                                                 test: 22~hrs, 0.8~k pairs} \\
        \hline
        \makecell[c]{data creation\\constraint} & \makecell[c]{\texttt{count} inter-category audio\\ \emph{SpatialDist} intra-category\\and require temporder}\\
        \hline
        \makecell[c]{audio\\diversity} & \makecell[c]{one event $\rightarrow$ multi-audios;\\
                                                    seed audio $\rightarrow$ multi time len;\\
                                  seed audios various start time \\}\\
        \hline
        \makecell[c]{text prompts \\diversity} & \makecell[c]{GPT-4 augmented prompts;\\
                                  one template $\rightarrow$ multi-events.}\\
        \bottomrule
        \end{tabular}
    \caption{\emph{RiTTA} benchmark highlights.}
    \label{tab:ritta_datacreate}
\end{table}

\section{Key Findings of TTA models on RiTTA Benchmark}
\label{key_findings}

The quantitative evaluation results~(mAMSR score) w.r.t the four main relation categories are presented in Table~\ref{tab:results-fourcates}. We observe that both general and relation-aware evaluations show better performance on \emph{Temporal Order} and \emph{Compositionality} compared to \emph{Count} and \emph{Spatial Distance}. This suggests that the \emph{Count} and \emph{Spatial Distance} relations pose significant challenges for TTA tasks. Additionally, we visualize the detailed relation aware evaluation results for the 11 sub-relations, highlighting the top three performing methods AudioLDM~2~\cite{audioldm2}, Tango~\cite{ghosal2023tango}, and Tango~2~\cite{ghosal2023tango2}, in Fig.~\ref{fig:exp_radar_chart}. We can observe that all the three methods 1. achieve exceedingly high presence score on \texttt{Not} relation, which is expected since a high \textbf{Presence} score~(Subfig.~A) can be easily obtained by simply not generating the specified audio event. 2. perform well in modeling \texttt{And} relation~(Subfig.~B)~(then \texttt{<equaldist>} and the three relations in \emph{Temporal Order}); 3. exhibit strength in generating concise audios particularly for \texttt{Not} relation~(Subfig.~C). Overall, all the three methods excel in modeling \texttt{And} relation and then the three sub-relations in \emph{Temporal Order}, which is also reflected by the result in Table~\ref{tab:results-fourcates}. The key findings from the relation-aware benchmarking are summarized in the Table~\ref{tab:benchmark_findings}.

The key findings of TTA models is summarized in Table~\ref{tab:benchmark_findings}.

\begin{table}[h]
\centering
\begin{tabular}{l}
\toprule
\textbf{1.} generation eval. contradicts with RiTTA eval.\\
\textbf{2.} \emph{TemOrder}/\emph{Compos} better than \emph{Count}/\emph{SpatDist}\\
\textbf{3.} event presence in \texttt{Not} is the highest;\\
\textbf{4.} relation correctness in \texttt{And} is the highest;\\
\textbf{5.} parsimony score in \texttt{Not} is the highest;\\
\textbf{6.} event presence accuracy rate is below $1\%$; \\
\textbf{7.} relation correctness accuracy rate is below $1\%$;\\
\textbf{8.} An average of 2 redundant audio events;\\
\bottomrule
\end{tabular}
\caption{Key findings from experiments of TTA models on our RiTTA benchmark.}
\label{tab:benchmark_findings}
\end{table}

\section{Conclusion and Future Works}
\label{sec:appen:conclusion}

Complex relationships within audio bring the world to life. While text-to-audio~(TTA) generation models have made remarkable progress in generating high-fidelity audio with fine-grained context understanding, they often fall short in capturing the relational aspect of audio events in real-world. The world around us is composed of interconnected audio events, where audio event rarely occurs in isolation. Simply generating single sound sources is insufficient for producing realistic audio that reflects the richness of the world.

To analyze the capabilities of current state-of-the-art TTA generative models, we first conduct a systematic study of these models in audio event relation modeling. We introduce a benchmark for this task by creating a comprehensive relational corpus covering all potential relations in the real-world scenarios. Further, we propose new evaluation metric framework to assess audio event relation modeling from various perspectives. Additionally, we propose a finetuning strategy to boost existing models' ability in modelling audio events relation, and we show improvement across all relation metrics. Finally, we will release both the dataset and the code for the evaluation metrics, which will be useful for future research in this domain.

Going forward, our work provides a unique research opportunity to bring the world to life by exploring ways to generate long-term audio events to acoustically understand the physical world. Further, understanding the successes and failures of these models in generating such complex audio events is another promising research direction. This analysis could lead to further improvements in TTA models and their applications in areas such as virtual reality, cinema and immersive media.

\end{document}